\documentclass{article}

\usepackage{arxiv}

\usepackage[utf8]{inputenc} 
\usepackage[T1]{fontenc}    
\usepackage[pdfborder={0 0 0}]{hyperref}       
\usepackage{url}            
\usepackage{booktabs}       
\usepackage{amsfonts}       
\usepackage{nicefrac}       
\usepackage{microtype}      
\usepackage{lipsum}		
\usepackage{graphicx}
\usepackage{natbib}
\usepackage{doi}
\title{ChatGPT outperforms crowd-workers\\for text-annotation tasks}

\date{Published in the \emph{Proceedings of the National Academy of Sciences}\\\url{https://www.pnas.org/doi/10.1073/pnas.2305016120}}	

\author{
{Fabrizio Gilardi}\thanks{\normalsize{Corresponding author (\url{https://fabriziogilardi.org/}).}}\\
University of Zurich\\
Zurich, Switzerland\\
\And
{Meysam Alizadeh}\\
University of Zurich\\
Zurich, Switzerland\\
\And
{Maël Kubli}\\
University of Zurich\\
Zurich, Switzerland\\
}

\renewcommand{\headeright}{}
\renewcommand{\undertitle}{}
\renewcommand{\shorttitle}{ChatGPT outperforms crowd-workers for text-annotation tasks}

\hypersetup{
pdftitle={ChatGPT outperforms crowd-workers for text-annotation tasks},
pdfsubject={},
pdfauthor={Fabrizio Gilardi, Meysam Alizadeh, Maël Kubli},
pdfkeywords={ChatGPT, LLMs, Large Language Models, NLP, Text Annotation},
}

\begin{document}

\maketitle

\begin{abstract}
Many NLP applications require manual text annotations for a variety of tasks, notably to train classifiers or evaluate the performance of unsupervised models. Depending on the size and degree of complexity, the tasks may be conducted by crowd-workers on platforms such as MTurk as well as trained annotators, such as research assistants. Using four samples of tweets and news articles (n = 6,183), we show that ChatGPT outperforms crowd-workers for several annotation tasks, including relevance, stance, topics, and frame detection. Across the four datasets, the zero-shot accuracy of ChatGPT exceeds that of crowd-workers by about 25 percentage points on average, while ChatGPT's intercoder agreement exceeds that of both crowd-workers and trained annotators for all tasks. Moreover, the per-annotation cost of ChatGPT is less than \$0.003---about thirty times cheaper than MTurk. These results demonstrate the potential of large language models to drastically increase the efficiency of text classification. 
\end{abstract}


\section{Introduction}

Many NLP applications require high-quality labeled data, notably to train classifiers or evaluate the performance of unsupervised models. For example, researchers often aim to filter noisy social media data for relevance, assign texts to different topics or conceptual categories, or measure their sentiment or stance. Regardless of the specific approach used for these tasks (supervised, semi-supervised, or unsupervised), labeled data are needed to build a training set or a gold standard against which performance can be assessed. Such data may be available for high-level tasks such as semantic evaluation \citep{Emerson:2022aa}. More typically, however, researchers have to conduct original annotations to ensure that the labels match their conceptual categories \citep{Benoit:2016rt}. Until recently, two main strategies were available. First, researchers can recruit and train coders, such as research assistants. Second, they can rely on crowd-workers on platforms such as Amazon Mechanical Turk (MTurk). Often, these two strategies are used in combination: trained annotators create a relatively small gold-standard dataset, and crowd-workers are employed to increase the volume of labeled data. Trained annotators tend to produce high-quality data, but involve significant costs. Crowd workers are a much cheaper and more flexible option, but the quality may be insufficient, particularly for complex tasks and languages other than English. Moreover, there have been concerns that MTurk data quality has decreased \citep{Chmielewski:2020aa}, while alternative platforms such as CrowdFlower and FigureEight are no longer practicable options for academic research since they were acquired by Appen, a company that is focused on a business market.

This paper explores the potential of large language models (LLMs) for text annotation tasks, with a focus on ChatGPT, which was released in November 2022. It demonstrates that zero-shot ChatGPT classifications (that is, without any additional training) outperform MTurk annotations, at a fraction of the cost. LLMs have been shown to perform very well for a wide range of purposes, including ideological scaling \citep{Wu:2023aa}, the classification of legislative proposals \citep{Nay:2023aa}, the resolution of cognitive psychology tasks \citep{binz2023using}, and the simulation of human samples for survey research \citep{Argyle:2023aa}. While a few studies suggested that ChatGPT might perform text annotation tasks of the kinds we have described \citep{kuzman2023chatgpt, huang2023chatgpt}, to the best of our knowledge our work is the first systematic evaluation. Our analysis relies on a sample of 6,183 documents, including tweets and news articles that we collected for a previous study \citep{alizadeh2022content} as well as a new sample of tweets posted in 2023. In our previous study, the texts were labeled by trained annotators (research assistants) for five different tasks: relevance, stance, topics, and two kinds of frame detection. Using the same codebooks that we developed to instruct our research assistants, we submitted the tasks to ChatGPT as zero-shot classifications, as well as to crowd-workers on MTurk. We then evaluated the performance of ChatGPT against two benchmarks: (i) its accuracy, relative to that of crowd-workers, and (ii) its intercoder agreement, relative to that of crowd workers as well as of our trained annotators. We find that across the four datasets, ChatGPT's zero-shot accuracy is higher than that of MTurk for most tasks. For all tasks, ChatGPT’s intercoder agreement exceeds that of both MTurk and trained annotators. Moreover, ChatGPT is significantly cheaper than MTurk. ChatGPT's per-annotation cost is about \$0.003, or a third of a cent---about thirty times cheaper than MTurk, with higher quality. At this cost, it might potentially be possible to annotate entire samples, or to create large training sets for supervised learning. While further research is needed to better understand how ChatGPT and other LLMs perform in a broader range of contexts, these results demonstrate their potential to transform how researchers conduct data annotations, and to disrupt parts of the business model of platforms such as MTurk. 

\section{Results}

We use four datasets ($n = 6,183$) including tweets and news articles that we collected and annotated manually for a previous study on the discourse around content moderation \citep{alizadeh2022content}, as well as a new sample of tweets posted in 2023 to address the concern that ChatGPT might be relying on memorization for texts potentially included in the model's training dataset. We relied on trained annotators (research assistants) to construct a gold standard for six conceptual categories: relevance of tweets for the content moderation issue (relevant/irrelevant); relevance of tweets for political issues (relevant/irrelevant); stance regarding Section 230, a key part of US internet legislation (keep/repeal/neutral); topic identification (six classes); a first set of frames (content moderation as a problem, as a solution, or neutral); and a second set of frames (fourteen classes). We then performed these exact same classifications with ChatGPT and with crowd-workers recruited on MTurk, using the same codebook we developed for our research assistants (see SI Appendix). For ChatGPT, we conducted four sets of annotations. To explore the effect of ChatGPT's temperature parameter, which controls the degree of randomness of the output, we conducted the annotations with the default value of 1 as well as with a value of 0.2, which implies less randomness. For each temperature value, we conducted two sets of annotations to compute ChatGPT's intercoder agreement. For MTurk, we aimed to select high-quality crowd-workers, notably by filtering for workers who are classified as ``MTurk Masters'' by Amazon, who have an approval rate of over 90\%, and who are located in the US. Our procedures are described more in detail in the \emph{Materials and Methods} section.

\begin{figure}
\centering
\includegraphics[width=0.9\linewidth]{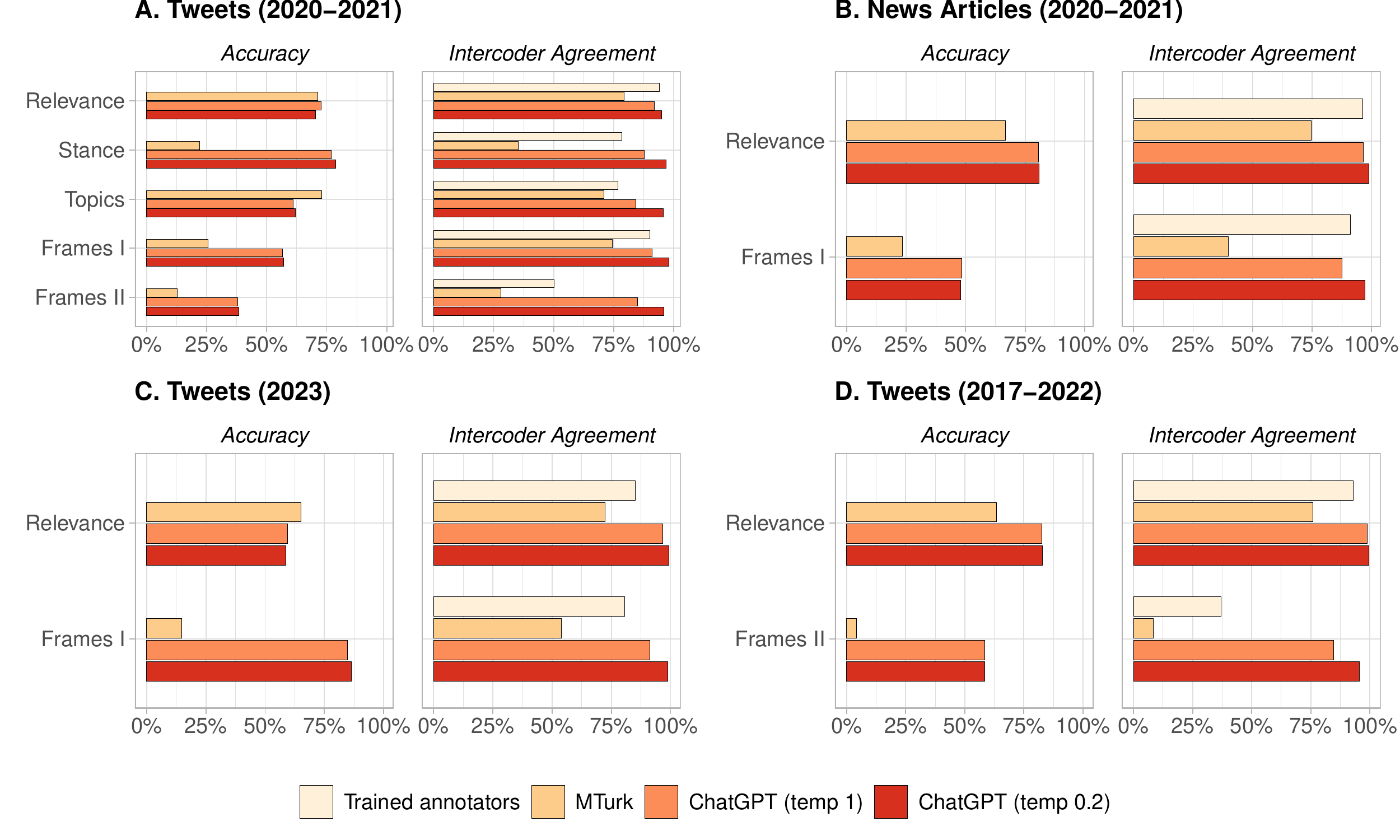}
\caption{ChatGPT zero-shot text annotation performance in four datasets, compared to MTurk and trained annotators. ChatGPT's accuracy outperforms that of MTurk for most tasks. ChatGPT's intercoder agreement outperforms that of both MTurk and trained annotators in all tasks. Accuracy means agreement with the trained annotators.}
\label{fig:performance}
\end{figure}

Across the four datasets, we report ChatGPT's zero-shot performance for two different metrics: accuracy and intercoder agreement (Figure \ref{fig:performance}). Accuracy is measured as the percentage of correct annotations (using our trained annotators as a benchmark), while intercoder agreement is computed as the percentage of tweets that were assigned the same label by two different annotators (research assistant, crowd-workers, or ChatGPT runs). Regarding accuracy, Figure \ref{fig:performance} shows that ChatGPT outperforms MTurk for most tasks across the four datasets. On average, ChatGPT's accuracy exceeds that of MTurk by about 25 percentage points. Moreover, ChatGPT demonstrates adequate accuracy overall, considering the challenging tasks, number of classes, and zero-shot annotations. Accuracy rates for relevance tasks, with two classes (relevant/irrelevant) are 70\% for content moderation tweets, 81\% for content moderation news articles, 83\% for US Congress tweets, and 59\% for 2023 content moderation tweets. In the 2023 sample, ChatGPT performed much better than MTurk in the second task but struggled with misclassifying tweets about specific user suspensions in the relevance task due to a lack of examples in the prompt. While these findings do not suggest that memorization is a major issue, they underscore the importance of high-quality prompts.

Regarding intercoder agreement, Figure \ref{fig:performance} shows that ChatGPT's performance is very high. On average, intercoder agreement is about 56\% for MTurk, 79\% for trained annotators, 91\% for ChatGPT with temperature = 1, and 97\% for ChatGPT with temperature = 0.2. The correlation between intercoder agreement and accuracy is positive (Pearson's r = 0.36). This suggests that a lower temperature value may be preferable for annotation tasks, as it seems to increase consistency without decreasing accuracy.

We underscore that the test to which we subjected ChatGPT is hard. Our tasks were originally conducted in the context of a previous study \citep{alizadeh2022content}, and required considerable resources. We developed most of the conceptual categories for our particular research purposes. Moreover, some of the tasks involve a large number of classes and exhibit lower levels of intercoder agreement, which indicates a higher degree of annotation difficulty \citep{Bayerl:2011aa}. ChatGPT's accuracy is positively correlated with the intercoder agreement of trained annotators (Pearson's r $= 0.46$), suggesting better performance for easier tasks. Conversely, ChatGPT's outperformance of MTurk is negatively correlated with the intercoder agreement of trained annotators (Pearson's r $= -0.37$), potentially indicating stronger overperformance for more complex tasks.

We conclude that ChatGPT's performance is impressive, particularly considering that its annotations are zero-shot.

\section{Discussion}

This paper demonstrates the potential of LLMs to transform text-annotation procedures for a variety of tasks common to many research projects. The evidence is consistent across different types of texts and time periods. It strongly suggests that ChatGPT may already be a superior approach compared to crowd-annotations on platforms such as MTurk. At the very least, the findings demonstrate the importance of studying the text-annotation properties and capabilities of LLMs more in depth. The following questions seem particularly promising: (i) performance across multiple languages; (ii) implementation of few-shot learning; (iii) construction of semi-automated data labeling systems in which a model learns from human annotations and then recommends labeling procedures \citep{desmond2021semi}; (iv) using chain of thought prompting and other strategies to increase the performance of zero-shot reasoning \citep{kojima2022large}; and (v) comparison across different types of LLMs. 


\section{Materials and Methods}

\subsection{Datasets}

The analysis relies on four datasets: (i) a random sample of 2,382 tweets drawn from a dataset of 2.6 million tweets on content moderation posted from January 2020 to April 2021; (ii) a random sample of 1,856 tweets posted by members of the US Congress from 2017 to 2022, drawn from a dataset of 20 million tweets; (iii) a random sample of 1,606 articles newspaper articles on content moderation published from January 2020 to April 2021, drawn from a dataset of 980k articles collected via LexisNexis. Sample size was determined by the number of texts needed to build a training set for a machine learning classifier. The fourth dataset (iv) replicated the data collection for (i), but for January 2023. It includes a random sample of 500 tweets (of which 339 were in English) drawn from a dataset of 1.3 million tweets. 

\subsection{Annotation Tasks}\label{sec:annotation}

We implemented several annotation tasks: (1) \textit{relevance}: whether a tweet is about content moderation or, in a separate task, about politics; (2) \textit{topic detection}: whether a tweet is about a set of six pre-defined topics (i.e. Section 230, Trump Ban, Complaint, Platform Policies, Twitter Support, and others); (3) \textit{stance detection}: whether a tweet is in favor of, against, or neutral about repealing Section 230 (a piece of US legislation central to content moderation); (4) \textit{general frame detection}: whether a tweet contains a set of two opposing frames  (``problem' and ``solution''). The solution frame describes tweets framing content moderation as a solution to other issues (e.g., hate speech). The problem frame describes tweets framing content moderation as a problem on its own as well as to other issues (e.g., free speech); (5) \textit{policy frame detection}: whether a tweet contains a set of fourteen policy frames proposed in \citep{card2015media}. The full text of instruction for the five annotation tasks is presented in SI Appendix. We used the exact same wordings for ChatGPT and MTurk.

\subsection{Trained annotators}

We trained three political science students to conduct the annotation tasks. For each task, they were given the same set of instructions described above and detailed in SI Appendix. The coders annotated the tweets independently task by task.

\subsection{Crowd-workers}

We employed MTurk workers to perform the same set of tasks as trained annotators and ChatGPT, using the same set of instructions (SI Appendix). To ensure annotation quality, we restricted access to the tasks to workers who are classified as ``MTurk Masters'' by Amazon, who have a HIT (Human Intelligence Task) approval rate greater than 90\% with at least 50 approved HITs, and who are located in the US. Moreover, we ensured that no worker could annotate more than 20\% of the tweets for a given task. As with the trained human annotators, each tweet was annotated by two different crowd-workers.  

\subsection{ChatGPT}

We used the ChatGPT API with the `gpt-3.5-turbo'. The annotations were conducted between March 9-20 and April 27-May 4, 2023. For each task, we prompted ChatGPT with the corresponding annotation instruction text (SI Appendix). We intentionally avoided adding any ChatGPT-specific prompts to ensure comparability between ChatGPT and MTurk crowd-workers. After testing several variations, we decided to feed tweets one by one to ChatGPT using the following prompt: ``Here's the tweet I picked, please label it as [Task Specific Instruction (e.g. `one of the topics in the instruction')].'' We set the \textit{temperature} parameter at 1 (default value) and 0.2 (which makes the output more deterministic; higher values make the output more random). For each temperature setting, we collected two responses from ChatGPT to compute the intercoder agreement. That is, we collected four ChatGPT responses for each tweet. We created a new chat session for every tweet to ensure that the ChatGPT results are not influenced by the history of annotations.

\subsection{Evaluation Metrics}

First, we computed average accuracy (i.e. percentage of correct predictions), that is, the number of correctly classified instances over the total number of cases to be classified, using trained human annotations as our gold standard and considering only texts that both annotators agreed upon. Second, intercoder agreement refers to the percentage of instances for which both annotators in a given group report the same class.

\subsection{Data Availability}

Replication materials are available at the Harvard Dataverse, \url{https://doi.org/10.7910/DVN/PQYF6M}.


\section*{Acknowledgments}

This project received funding from the European Research Council (ERC) under the European Union's Horizon 2020 research and innovation program (grant agreement nr. 883121). We thank Fabio Melliger, Paula Moser, and Sophie van IJzendoorn for excellent research assistance.

\bibliographystyle{apsr}
\bibliography{pnas-sample.bib}

\newpage
\appendix
\pagenumbering{arabic}
\renewcommand{\thesection}{S\arabic{section}}

\section{Annotation Codebooks}\label{sec:prompt}

Not all of the annotations described in these codebooks were conducted for every dataset in our study. First, the manual annotations we use as a benchmark were performed in a previous study, except for the new 2023 sample, which was specifically annotated for this current study. Second, certain annotation tasks are not applicable to all datasets. For instance, stance analysis, problem/solution, and topic modeling were not suitable for analyzing tweets from US Congress members. This is because these tweets cover a wide range of issues and topics, unlike content moderation topics, which are more focused. For news articles, our attempts at human annotation for stance, topic, and policy frames were not successful. This was because the articles primarily revolved around platform policies, actions, and criticisms thereof. 

\subsection{Background on content moderation (to be used for all tasks except the tweets from  US Congressmembers)}
For this task, you will be asked to annotate a sample of tweets about content moderation. Before describing the task, we explain what we mean by “content moderation”. 

“Content moderation” refers to the practice of screening and monitoring content posted by users on social media sites to determine if the content should be published or not, based on specific rules and guidelines. Every time someone posts something on a platform like Facebook or Twitter, that piece of content goes through a review process (“content moderation”) to ensure that it is not illegal, hateful or inappropriate and that it complies with the rules of the site. When that is not the case, that piece of content can be removed, flagged, labeled as or ‘disputed.’ 

Deciding what should be allowed on social media is not always easy. For example, many sites ban child pornography and terrorist content as it is illegal. However, things are less clear when it comes to content about the safety of vaccines or politics, for example. Even when people agree that some content should be blocked, they do not always agree about the best way to do so, how effective it is, and who should do it (the government or private companies, human moderators, or artificial intelligence).

\subsection{Background on political tweets (to be used for tweets by the US Congress members)}
For this task, you will be asked to annotate a sample of tweets to determine if they include political content or not. For the purposes of this task, tweets are “relevant” if they include political content, and “irrelevant” if they do not. Before describing the task, we explain what we mean by “political content”.

“Political content” refers to any tweets that pertain to politics or government policies at the local, national, or international level. This can include tweets that discuss political figures, events, or issues, as well as tweets that use political language or hashtags. To determine if tweets include political content or not, consider several factors, such as the use of political keywords or hashtags,  the mention of political figures or events, the inclusion of links to news articles or other political sources, and the overall tone and sentiment of the tweet, which may indicate whether it is conveying a political message or viewpoint.

\subsection{Task 1: Relevance (Content Moderation)}
For each tweet in the sample, follow these instructions:

\begin{enumerate}
   \item Carefully read the text of the tweet, paying close attention to details.
   \item Classify the tweet as either relevant (1) or irrelevant (0) 
\end{enumerate}
    
Tweets should be coded as RELEVANT when they directly relate to content moderation, as defined above. This includes tweets that discuss: social media platforms’ content moderation rules and practices, governments’ regulation of online content moderation, and/or mild forms of content moderation like flagging.

Tweets should be coded as IRRELEVANT if they do not refer to content moderation, as defined above, or if they are themselves examples of moderated content. This would include, for example, a Tweet by Donald Trump that Twitter has labeled as “disputed”, a tweet claiming that something is false, or a tweet containing sensitive content. Such tweets might be subject to content moderation, but are not discussing content moderation. Therefore, they should be coded as irrelevant for our purposes.

\subsection{Task 2: Relevance (Political Content)}
For each tweet in the sample, follow these instructions:

\begin{enumerate}
   \item Carefully read the text of the tweet, paying close attention to details.
   \item Classify the tweet as either relevant (1) or irrelevant (0) 
\end{enumerate}

Tweets should be coded as RELEVANT if they include POLITICAL CONTENT, as defined above. 
Tweets should be coded as IRRELEVANT if they do NOT include POLITICAL CONTENT, as defined above.

\subsection{Task 3: Problem/Solution Frames}
Content moderation can be seen from two different perspectives:

\begin{itemize}
   \item Content moderation can be seen as a PROBLEM; for example, as a restriction of free speech
   \item Content moderation can be seen as a SOLUTION; for example, as a protection from harmful speech 
\end{itemize}

For each tweet in the sample, follow these instructions:

\begin{enumerate}
   \item Carefully read the text of the tweet, paying close attention to details.
   \item Classify the tweet as describing content moderation as a problem, as a solution, or neither.
\end{enumerate}

Tweets should be classified as describing content moderation as a PROBLEM if they emphasize negative effects of content moderation, such as restrictions to free speech, or the biases that can emerge from decisions regarding what users are allowed to post.

Tweets should be classified as describing content moderation as a SOLUTION if they emphasize positive effects of content moderation, such as protecting users from various kinds of harmful content, including hate speech, misinformation, illegal adult content, or spam. 

Tweets should be classified as describing content moderation as NEUTRAL if they do not emphasize possible negative or positive effects of content moderation, for example if they simply report on the content moderation activity of social media platforms without linking them to potential advantages or disadvantages for users or stakeholders.

\subsection{Task 4: Policy Frames (Content Moderation)}

Content moderation, as described above, can be linked to various other topics, such as health, crime, or equality. 

For each tweet in the sample, follow these instructions:

\begin{enumerate}
   \item Carefully read the text of the tweet, paying close attention to details.
  \item Classify the tweet into one of the topics defined below.
\end{enumerate}

The topics are defined as follows:

\begin{itemize}
   \item ECONOMY: The costs, benefits, or monetary/financial implications of the issue (to an individual, family, community, or to the economy as a whole).
   \item Capacity and resources: The lack of or availability of physical, geographical, spatial, human, and financial resources, or the capacity of existing systems and resources to implement or carry out policy goals.
   \item MORALITY: Any perspective—or policy objective or action (including proposed action)that is compelled by religious doctrine or interpretation, duty, honor, righteousness or any other sense of ethics or social responsibility.
   \item FAIRNESS AND EQUALITY: Equality or inequality with which laws, punishment, rewards, and resources are applied or distributed among individuals or groups. Also the balance between the rights or interests of one individual or group compared to another individual or group.
   \item CONSTITUTIONALITY AND JURISPRUDENCE: The constraints imposed on or freedoms granted to individuals, government, and corporations via the Constitution, Bill of Rights and other amendments, or judicial interpretation. This deals specifically with the authority of government to regulate, and the authority of individuals/corporations to act independently of government.
   \item POLICY PRESCRIPTION AND EVALUATION: Particular policies proposed for addressing an identified problem, and figuring out if certain policies will work, or if existing policies are effective.
   \item LAW AND ORDER, CRIME AND JUSTICE: Specific policies in practice and their enforcement, incentives, and implications. Includes stories about enforcement and interpretation of laws by individuals and law enforcement, breaking laws, loopholes, fines, sentencing and punishment. Increases or reductions in crime.
   \item SECURITY AND DEFENSE: Security, threats to security, and protection of one’s person, family, in-group, nation, etc. Generally an action or a call to action that can be taken to protect the welfare of a person, group, nation sometimes from a not yet manifested threat.
   \item HEALTH AND SAFETY: Health care access and effectiveness, illness, disease, sanitation, obesity, mental health effects, prevention of or perpetuation of gun violence, infrastructure and building safety.
   \item QUALITY OF LIFE: The effects of a policy on individuals’ wealth, mobility, access to resources, happiness, social structures, ease of day-to-day routines, quality of community life, etc.
   \item CULTURAL IDENTITY: The social norms, trends, values and customs constituting culture(s), as they relate to a specific policy issue.
   \item PUBLIC OPINION: References to general social attitudes, polling and demographic information, as well as implied or actual consequences of diverging from or “getting ahead of” public opinion or polls.
   \item POLITICAL: Any political considerations surrounding an issue. Issue actions or efforts or stances that are political, such as partisan filibusters, lobbyist involvement, bipartisan efforts, deal-making and vote trading, appealing to one's base, mentions of political maneuvering. Explicit statements that a policy issue is good or bad for a particular political party.
   \item EXTERNAL REGULATION AND REPUTATION: The United States’ external relations with another nation; the external relations of one state with another; or relations between groups. This includes trade agreements and outcomes, comparisons of policy outcomes or desired policy outcomes.
   \item OTHER: Any topic that does not fit into the above categories.
\end{itemize}

\subsection{Task 5: Policy Frames (Political Content)}

Political content, as described above, can be linked to various other topics, such as health, crime, or equality. 

For each tweet in the sample, follow these instructions:

\begin{enumerate}
   \item Carefully read the text of the tweet, paying close attention to details.
   \item Classify the tweet into one of the topics defined below.
\end{enumerate}

The topics are defined as follows:

\begin{itemize}
   \item ECONOMY: The costs, benefits, or monetary/financial implications of the issue (to an individual, family, community, or to the economy as a whole).
   \item Capacity and resources: The lack of or availability of physical, geographical, spatial, human, and financial resources, or the capacity of existing systems and resources to implement or carry out policy goals.
   \item MORALITY: Any perspective—or policy objective or action (including proposed action)that is compelled by religious doctrine or interpretation, duty, honor, righteousness or any other sense of ethics or social responsibility.
   \item FAIRNESS AND EQUALITY: Equality or inequality with which laws, punishment, rewards, and resources are applied or distributed among individuals or groups. Also the balance between the rights or interests of one individual or group compared to another individual or group.
   \item CONSTITUTIONALITY AND JURISPRUDENCE: The constraints imposed on or freedoms granted to individuals, government, and corporations via the Constitution, Bill of Rights and other amendments, or judicial interpretation. This deals specifically with the authority of government to regulate, and the authority of individuals/corporations to act independently of government.
   \item POLICY PRESCRIPTION AND EVALUATION: Particular policies proposed for addressing an identified problem, and figuring out if certain policies will work, or if existing policies are effective.
   \item LAW AND ORDER, CRIME AND JUSTICE: Specific policies in practice and their enforcement, incentives, and implications. Includes stories about enforcement and interpretation of laws by individuals and law enforcement, breaking laws, loopholes, fines, sentencing and punishment. Increases or reductions in crime.
   \item SECURITY AND DEFENSE: Security, threats to security, and protection of one’s person, family, in-group, nation, etc. Generally an action or a call to action that can be taken to protect the welfare of a person, group, nation sometimes from a not yet manifested threat.
   \item HEALTH AND SAFETY: Health care access and effectiveness, illness, disease, sanitation, obesity, mental health effects, prevention of or perpetuation of gun violence, infrastructure and building safety.
   \item QUALITY OF LIFE: The effects of a policy on individuals’ wealth, mobility, access to resources, happiness, social structures, ease of day-to-day routines, quality of community life, etc.
   \item CULTURAL IDENTITY: The social norms, trends, values and customs constituting culture(s), as they relate to a specific policy issue.
   \item PUBLIC OPINION: References to general social attitudes, polling and demographic information, as well as implied or actual consequences of diverging from or “getting ahead of” public opinion or polls.
   \item POLITICAL: Any political considerations surrounding an issue. Issue actions or efforts or stances that are political, such as partisan filibusters, lobbyist involvement, bipartisan efforts, deal-making and vote trading, appealing to one's base, mentions of political maneuvering. Explicit statements that a policy issue is good or bad for a particular political party.
   \item EXTERNAL REGULATION AND REPUTATION: The United States’ external relations with another nation; the external relations of one state with another; or relations between groups. This includes trade agreements and outcomes, comparisons of policy outcomes or desired policy outcomes.
   \item OTHER: Any topic that does not fit into the above categories.
\end{itemize}

\subsection{Task 6: Stance Detection}

In the context of content moderation, Section 230 is a law in the United States that protects websites and other online platforms from being held legally responsible for the content posted by their users. This means that if someone posts something illegal or harmful on a website, the website itself cannot be sued for allowing it to be posted. However, websites can still choose to moderate content and remove anything that violates their own policies.

For each tweet in the sample, follow these instructions:

\begin{enumerate}
   \item Carefully read the text of the tweet, paying close attention to details.
   \item Classify the tweet as having a positive stance towards Section 230, a negative stance, or a neutral stance.
\end{enumerate}

\subsection{Task 7: Topic Detection}

Tweets about content moderation may also discuss other related topics, such as:

\begin{enumerate}
   \item Section 230, which is a law in the United States that protects websites and other online platforms from being held legally responsible for the content posted by their users (SECTION 230).
   \item The decision by many social media platforms, such as Twitter and Facebook, to suspend Donald Trump’s account (TRUMP BAN).
   \item Requests directed to Twitter’s support account or help center (TWITTER SUPPORT).
   \item Social media platforms’ policies and practices, such as community guidelines or terms of service (PLATFORM POLICIES).
   \item Complaints about platform’s policy and practices in deplatforming and content moderation or suggestions to suspend particular accounts, or complaints about accounts being suspended or reported  (COMPLAINTS).
   \item If a text is not about the SECTION 230, COMPLAINTS, TRUMP BAN, TWITTER SUPPORT, and PLATFORM POLICIES, then it should be classified in OTHER class (OTHER).
\end{enumerate}

For each tweet in the sample, follow these instructions:

\begin{enumerate}
   \item Carefully read the text of the tweet, paying close attention to details.
   \item Please classify the following text according to topic (defined by function of the text, author’s purpose and form of the text). You can choose from the following classes: SECTION 230, TRUMP BAN, COMPLAINTS, TWITTER SUPPORT, PLATFORM POLICIES, and OTHER
\end{enumerate}

\end{document}